\documentclass[runningheads]{llncs}

\usepackage{eccv}

\usepackage{eccvabbrv}

\usepackage{graphicx}
\usepackage{booktabs}

\usepackage[accsupp]{axessibility}  

\usepackage{multirow}

\usepackage{lineno}
\newcommand{\rev}[1]{\textcolor{black}{#1}}

\usepackage[pagebackref,breaklinks,colorlinks,citecolor=eccvblue]{hyperref}

\usepackage{orcidlink}

\begin{document}

\title{MST-KD: Multiple Specialized Teachers Knowledge Distillation for Fair Face Recognition} 

\titlerunning{MST-KD: Multiple Specialized Teachers KD for Fair FR}

\author{Eduarda Caldeira\inst{1, 2} \and
Jaime S. Cardoso\inst{1,2} \and
Ana F. Sequeira\inst{1,2} \and \\Pedro C. Neto\inst{1,2}}

\authorrunning{E.~Caldeira et al.}

\institute{INESC TEC, Porto, Portugal \and University of Porto, Porto, Portugal\\
\email{melcaldeira22@gmail.com}}

\maketitle

\begin{abstract}
  As in school, one teacher to cover all subjects is insufficient to distill equally robust information to a student. Hence, each subject is taught by a highly specialised teacher. Following a similar philosophy, we propose a multiple specialized teacher framework to distill knowledge to a student network. In our approach, directed at face recognition use cases, we train four teachers on one specific ethnicity, leading to four highly specialized and biased teachers. Our strategy learns a project of these four teachers into a common space and distill that information to a student network. Our results highlighted increased performance and reduced bias for all our experiments. In addition, we further show that having biased/specialized teachers is crucial by showing that our approach achieves better results than when knowledge is distilled from four teachers trained on balanced datasets. Our approach represents a step forward to the understanding of the importance of ethnicity-specific features.
\end{abstract}

\section{Introduction} \label{sec:intro}

The development of Artificial Intelligence (AI) and deep learning (DL) has gained momentum over the past years. One concern that has been raised several times is related to the fairness of these algorithms, and how equally they perform on the different groups~\cite{melo2023synthesis}. In Face Recognition (FR), the nature, the environment and the risk of such decisions have led to an increased interest in mitigating any lack of fairness~\cite{neto2023compressed} as they handle human data. This means that despite the ability to create highly capable DL models for face recognition, one needs to take in consideration fairness aspects and the impact of deploying an unfair model~\cite{terhorst2020compr,neto2024beyond}.

In line with the performance improvements of DL-based face recognition systems, their opaqueness has also increased. This means that while the performance is improving, we lose the ability to infer the reasons behind a decision or to explain the reasoning of the model~\cite{neto2022explainable}. Inevitably leading to inability to directly understand if the model is discriminating against one or more demographic groups. Furthermore, one cannot easily understand which features are being learnt and if they relate more to a privileged group.  

Knowledge distillation (KD) figures as an interesting alternative to some of these problems. First, in a era where dataset retractions are getting more frequent due to ethical concerns~\cite{guo2016ms,DBLP:conf/fgr/CaoSXPZ18}, the ability to leverage the knowledge learnt by a teacher model to train a student model on a different dataset is of high relevance. In addition, if one is in the possession of a fair teacher, it can be used to mitigate biases originating from the training data. With the need for small models for deployment and the knowledge that smaller models trained from scratch lead to higher bias~\cite{neto2023compressed}, KD can be a solution to these biases~\cite{DBLP:conf/iccvw/LiuZSYL21, blakeney2021simon}, reducing the capacity gap associated with the student's lower complexity, or to address challenging scenarios, facilitating the student's learning with harder \cite{huber2021mask} or lower-quality \cite{ge2018low,ge2020efficient,boutros2022low} data. 

Despite usually being a straightforward task, using KD to train a student model is associated with some issues. Apart from the higher complexity of the training process and data considerations, the distillation itself is associated with a black-box behaviour that does not allow for an easy interpretation of which knowledge is being distilled. Specially if we consider a generalist teacher with no speciality, which means a teacher that is enforced to learn the highest number of common features between groups. In this sense, we hypothesize that having generalist teachers leads to lower performance and fairness. One can visualize this strategy as an educational system, where a student is provided with multiple professors/teachers highly specialized in a topic. Together, they contribute to define future of the student, and the student itself has the responsibility to learn from these multiple domains. Furthermore, this allows to trace, more accuratly, which teacher is distilling which information.

In this work, we use multiple specialized teachers to distill knowledge to a student network with reduced racial bias, resulting in the proposed multiple specialized teachers KD (MST-KD) framework. Each teacher is trained to distinguish identities from a specific ethnicity as each hyper-specialized model learns ethnicity-specific information that would not be easily gathered by a system 
addressing the more complex task of 
FR for all ethnicities. Distilling this knowledge to a single student 
might contribute to individually boost its performance for each ethnicity, 
resulting in increased global performance and racial fairness. 

We designed a set of experiments to showcase the impact of having hyper-specialised teachers. In that sense, we demonstrate that students trained with information distilled from specialized/biased teachers have higher performance metrics and better bias indicators in comparison to students trained with information from multiple balanced/generalist teachers. This presents itself as an intriguing behaviour, that indicates an advantage of learning ethnicity-specific information in comparison to jointly optimising for all ethnic groups. In this paper, we present the following contributions: 

\begin{itemize}
    \item A multi-teacher fusion approach that aggregates multiple N-dimensional spaces into a single N-dimensional space, allowing for KD; 
    \item We show that students that receive knowledge from hyper-specialised teachers achieve superior performance even if the teachers' individual performance is lower than that of balanced teachers;
    \item Our linear fusion allows for tracing the teacher representing the main source of the information distilled to the student;  
    \item An extensive analysis comprising multiple teacher and fusion backbones, and knowledge distillation with and without auxiliary classification losses.  
\end{itemize}


This document is organized as follows. Section \ref{sec:background} presents background information on bias and KD. The current literature on bias mitigation and KD in biometrics is explored in Section \ref{sec:sota}. Section \ref{sec:methodology} introduces this work's methodology and Section \ref{sec:results} presents a detailed analysis of the obtained results. The code used to perform the described experiments is publicly available at GitHub\footnote{\href{https://github.com/EduardaCaldeira/MST-KD}{https://github.com/EduardaCaldeira/MST-KD}}. Section \ref{sec:conclusion} summarizes the main conclusions and suggests possible future work directions. 

\section{Background} \label{sec:background}
This section presents relevant background information on bias (Section \ref{sec:back_bias}) and knowledge distillation (Section \ref{sec:back_KD}).

\subsection{Bias} \label{sec:back_bias}

The discrimination of specific demographic groups is and has always been a problem in society, resulting in unfavorable or even dangerous situations for the affected minorities. Even when the effects of prejudice are disregarded, bias is still involuntarily present in quotidian life, as humans are most likely to correctly identify people with demographic characteristics they contact with often, which usually correspond to their own \cite{robinson2020face}. This inadvertent bias can have serious consequences, potentially leading to life-altering events like wrongful convictions potentiated by incorrect witness testimonies \cite{innocenceproject2024race}.

DL methods are highly based on the human learning process and, thus, can also be biased. Bias in DL can be described as the presence of performance differences between sub-populations of the considered data \cite{robinson2020face}. When racial bias is considered, these sub-populations correspond to distinct ethnic groups. FR systems have already been proven to perform differently between ethnicities \cite{DBLP:journals/pami/HuangLLT20}, genders \cite{DBLP:conf/bmvc/AlbieroB20, DBLP:conf/icb/AlbieroZB20,DBLP:conf/icb/FuD22} and ages \cite{DBLP:conf/icb/DebN018}. These differences are typically linked to the challenge of differentiating between members of a given demographic group since identities belonging to the same population are typically easier to distinguish. The blind deployment of biased biometrics models is particularly problematic as it can enhance the demographic disparities verified in society due to humans' prejudiced beliefs \cite{caldeira2024model}. Hence, it is crucial to study bias mitigation strategies that allow for the development of fairer biometrics systems.

Differently from other tasks, FR is often evaluated on a verification scenario, where given a pair of images, the model predicts if they belong to the same identity. Hence, bias is frequently measured by comparing the difficulty of pairs from a certain demographic group versus other groups. This means that samples from one ethnicity are compared with samples from the same ethnicity, which means that ethnicity specific discriminative power is to be evaluated. 

The bias mitigation problem in DL can be approached from data-based or model-based perspectives. A substantial amount of FR datasets was gathered from the internet, resulting in unbalanced datasets due to the higher availability of Caucasian images on the internet. A relevant example is the CASIA-Webface dataset \cite{yi2014learning}, where 85\% of the samples belong to light-skin individuals \cite{wang2021meta}. 
Therefore, the creation of datasets with a balanced distribution of samples across demographic groups is the main goal of data-based bias mitigation strategies \cite{wang2019racial, robinson2020face, wang2021meta, karkkainen2021fairface}. However, using balanced training datasets is usually not enough to ensure fairness, as samples from distinct demographic groups constitute distinct challenges to the FR system. Regarding racial bias, it has been shown that these systems face higher difficulties when distinguishing non-Caucasian samples, which are inherently harder to classify \cite{wang2019racial}. Model-based strategies aim to tackle this issue through the correction of imbalances associated with the intrinsic complexity differences between demographics, either by fixing this complexity gap \cite{wang2021meta} or by detaching the FR process from the sensitive attributes \cite{xu2020investigating, franco2021learn}.

\subsection{Knowledge Distillation} \label{sec:back_KD}
KD techniques distill knowledge from a teacher model specialized in the proposed task to a student model that faces a knowledge gap. When this gap arises from the lower complexity of the student network, KD is used to guide its learning process through the more complex teacher, resulting in compressed networks with acceptable performance drops \cite{gou2021knowledge}. KD can also be used to develop networks with better performances in real-world scenarios, through the distillation of knowledge to a student trained on harder \cite{huber2021mask} or lower-quality \cite{ge2018low, ge2020efficient, boutros2022low} versions of the data processed by the teacher. 

Three main types of KD are usually defined according to the type of information distilled: response-based KD (RB-KD), feature-based KD (FB-KD), and relation KD (R-KD) \cite{caldeira2024model}. RB-KD \cite{ge2020efficient,zhao2023grouped} rely on the distillation of the teacher's soft probabilities, allowing the student to learn the teacher's predictions \cite{gou2021knowledge}. This restricts the usage of RB-KD to supervised approaches \cite{gou2021knowledge} while requiring access to the data used to train the teacher model to perform distillation, which is not always feasible in human-based applications such as FR \cite{kolf2023syper}. Furthermore, this strategy usually has a small impact on student's performance, as it does not consider important information encoded in the teacher's intermediate layers, especially when deep networks are considered \cite{gou2021knowledge}. FB-KD \cite{huber2021mask,boutros2022low,wu2020learning,wang2021teacher,liu2022coupleface} tackles these issues by distilling knowledge from one or more intermediate layers of the teacher model. This strategy results in more privacy-friendly since the distillation can be performed using unlabeled data, namely synthetic data \cite{kolf2023syper}. R-KD \cite{caldeira2023unveiling,liu2022coupleface,aslam2023privileged,boutros2022template,huang2022evaluation} softens the KD process by distilling relational information contained in the teacher's space. As only information regarding the teacher space's structural relations is transferred, this methodology removes the imposition of strictly trying to mimic the teacher space, which might be impossible when there is a wide knowledge gap between teacher and student \cite{aslam2023privileged}.

The KD process is usually incorporated in the student's loss function, which approximates its predictions (RB-KD), features (FB-KD) or correlations (R-KD) to the ones produced by the teacher for the same samples. When these samples' labels are available, a classification loss term can also be included, although its usage is not mandatory \cite{shahreza2023synthdistill}. In this work, FB-KD was used to distill knowledge from multiple teachers to a single student. To incorporate all the teachers' knowledge simultaneously in the KD process without spatial superposition, their spaces need to be adapted to a common multi-teacher space from where information can be directly distilled. Hence, the FB-KD loss can be written as:

\begin{equation}
    L_{FB_{KD}} = L_{KD}(\Psi(F_t), f_s),
    \label{eq:fb_simple}
\end{equation}
where $f_s$ represents the features that the student extracted from the layer involved in the distillation process, $F_t$ is the set of features extracted by all the considered teachers from the same layer, $\Psi$ is the adaptor function of the teachers' spaces and  $L_{KD}(.)$ is the selected KD loss function.
\section{State of the Art} \label{sec:sota}
This section summarizes the current literature on bias mitigation (Section \ref{sec:bias_SOTA}) and knowledge distillation (Section \ref{sec:KD_SOTA}). In Section \ref{sec:KD_bias}, these concepts meet to explain how KD methodologies have been used so far to address bias mitigation.

\subsection{Bias Mitigation} \label{sec:bias_SOTA}
To examine the variations in performance among demographic groups linked to bias, a test framework that enables comparisons based on the distinct sub-populations must be provided. The Racial Faces in the Wild (RFW) dataset \cite{wang2019racial} comprises an equal proportion of impostor-genuine pairs from African, Asian, Caucasian and Indian identities. FairFace \cite{karkkainen2021fairface} is a race-balanced dataset that is diverse in terms of race, age, expressions, head orientation, and photographic conditions; Balanced Faces in the Wild (BFW) \cite{robinson2020face} is balanced in terms of race, gender and race and gender simultaneously. Wang~\textit{et al.}~\cite{wang2021meta} developed two datasets that either follow the real-world skin-tone distribution (BUPT-Globalface) or present equal proportions of identities from the four considered ethnicities (BUPT-Balancedface).

Besides data-based bias mitigation strategies, model-based approaches have been proposed to increase the fairness of biometric models. Wang~\textit{et al.}~\cite{wang2021meta} evaluated the performance differences of the analyzed demographic groups considering their different complexity levels. To better align the model's feature space with the real data distribution, different adaptive margins were assigned to the evaluated skin tones, allowing harder skin tones to be assigned larger margins. Xu~\textit{et al.}~\cite{xu2020investigating} attempted to reduce bias in facial emotion recognition by either deleting or emphasizing sub-group attribute information throughout the recognition process. Franco~\textit{et al.}~\cite{franco2021learn} proved that having distributions independent of the attribute that can induce bias makes the model learn fairer representations of the data, resulting in improved fairness.

\subsection{Knowledge Distillation} \label{sec:KD_SOTA}
KD algorithms have previously aided the development of lighter models in biometrics, namely in FR. Boutros~\textit{et al.}~\cite{boutros2022pocketnet} achieved improved performance by condensing knowledge into a lightweight FR student through the use of a multistep KD (MS-KD) algorithm. Using a validation loss with a term that quantifies the computational cost of the defined model, Wang~\textit{et al.}~\cite{wang2021teacher} enabled the usage of Neural Architecture Search (NAS) \cite{zoph2016neural} to optimize the student's architecture. Neurons in the intermediate layers of a network usually encode both important attributes for the FR task, such as race and gender, and irrelevant information to this task. To penalize the impact of the latter in the KD process, Luo~\textit{et al.}~\cite{luo2016face} studied a KD loss term that does not include all the neurons of each layer.

However, KD algorithms are not restricted to model compression and acceleration tasks. Huber~\textit{et al.}~\cite{huber2021mask} bridged the accuracy gap of FR models in masked faces by distilling knowledge from a teacher trained on unmasked faces to a student trained on both masked and unmasked faces with remarkable results. KD was also used to bridge the performance gap between models trained on high and low-resolution images. Ge~\textit{et al.}~\cite{ge2018low} used FB-KD to distill knowledge from a teacher trained on high-resolution images to a student trained on their low-resolution versions. Caldeira~\textit{et al.}~\cite{caldeira2023unveiling} focused on the binary task of morphing attack detection, distilling distinct types of knowledge depending on each sample's class, achieving competitive results.

Distilling knowledge from a single teacher can improve student performance, but it is likely to lead to unfavorable consequences such as the distillation of biased predictions \cite{caldeira2024model}. This problem can be at least partly mitigated through the usage of several sources of information in the distillation procedure. The first multi-teacher KD methods proposed simple strategies to integrate the teachers' knowledge before distilling it to the student, namely attributing fixed weights to each teacher \cite{you2017learning, fukuda2017efficient, wu2019multi} to weight their importance levels differently during distillation. However, the teachers' relative capacity varies depending on the considered sample \cite{zhang2022confidence} and the layer from where knowledge is being distilled \cite{pham2023collaborative}, which cannot be trivially predicted, meaning that the relative importance of each teacher should be determined with finer granularity. Xu~\textit{et al.}~\cite{xu2023probabilistic} used KD to generate a student model from an ensemble of teachers, weighing each teacher's importance in a sample-wise manner. Pham~\textit{et al.}~\cite{pham2023collaborative} trained the teachers and the student simultaneously to include the teacher's weights to be learned during training, facilitating their optimization. Zhang~\textit{et al.}~\cite{zhang2022confidence} weighted each teacher' contribution in a sample-wise manner, considering their relative confidence level in the classification of each sample, given by the cross-entropy between their predictions and the ground truth labels. 

\subsection{Knowledge Distillation for Bias Mitigation} \label{sec:KD_bias}
KD has already been used to mitigate model bias. Liu~\textit{et al.}~\cite{DBLP:conf/iccvw/LiuZSYL21} proposed two strategies to extract a more uniform subsample of the considered training set, to ensure that useful knowledge is distilled during the KD procedure. These strategies outperformed simple KD in the evaluated scenarios, resulting in a significant improvement in fairness. As compressing a model can easily lead to increased bias levels \cite{caldeira2024model}, Blakeney~\textit{et al.}~\cite{blakeney2021simon} used KD to fine-tune a pruned network, proving that KD can effectively reduce the pruning-induced bias, resulting in fairer models. The present work proposes a framework that mitigates racial bias through the distillation of specialized knowledge on each ethnicity from different teachers. The student is trained on all ethnicity groups, benefiting from the ethnicity-specific knowledge provided by each teacher.
\section{Methodology} \label{sec:methodology}

In this section, the methodology followed in this work is presented. Section \ref{sec:data} describes the data used. In Section \ref{sec:teachers}, the proposed multi-teacher scheme is presented. Section \ref{sec:space_adaptors} details how the four teachers' spaces were adapted to a common multi-teacher space. Section \ref{sec:students} describes the student's architecture as well as the proposed knowledge distillation procedures. The baseline experiments are detailed in Section \ref{sec:baseline}. Finally, Section \ref{sec:exp_setup} describes the experimental setup followed during the networks' training, as well as the performance and fairness metrics used to evaluate the students.

\subsection{Data} \label{sec:data}
All the developed models were trained on the BUPT-Balancedface dataset \cite{wang2021meta}, which was specifically created to increase FR algorithms' fairness. This dataset contains 1.3 million images from African, Asian, Caucasian and Indian identities. Each ethnicity contributes 7k identities to the dataset, resulting in a race-balanced dataset with 28k identities.

Racial Faces in the Wild \cite{wang2019racial} was used to test the teacher and student networks. RFW is a benchmarking dataset for fair FR and contains the same percentage of genuine and impostor pairs from African, Asian, Caucasian and Indian identities. Each ethnicity contributes 3k identities with 6k image pairs for face verification, totaling 24k image pairs.

\subsection{Teachers} \label{sec:teachers}
\begin{figure}
    \centering
    \includegraphics[width=0.8\linewidth]{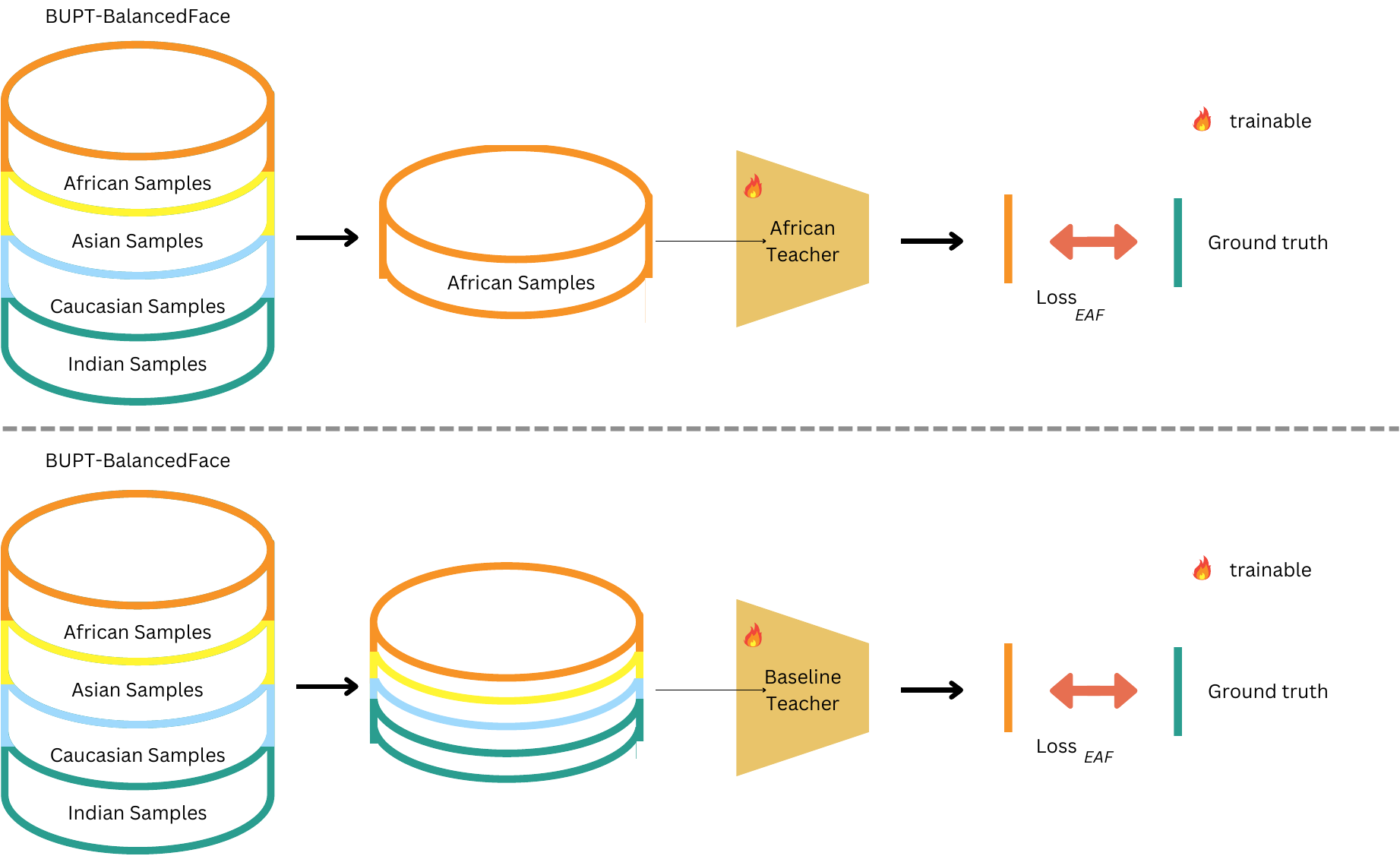}
    \caption{Different data splits performed before teacher training for our (top) and the baseline (bottom) multi-teacher approaches.}
    \label{fig:data_split}
\end{figure}

In this work, we use multiple teachers to distill knowledge to a student since specializing each teacher in distinguishing identities from a single ethnicity may facilitate the distillation of ethnicity-specific knowledge, boosting student performance and fairness. Hence, each of the four teachers was trained on the 7k identities of one of the considered ethnicities, as displayed in Figure \ref{fig:data_split}: African (T-Af), Asian (T-As), Caucasian (T-Ca) and Indian (T-In). These models are expected to perform better for identities from the ethnicity they were trained with since they are more familiar with their ethnicity-dependent attributes.

Each teacher consists of a backbone that processes the input samples, outputting 512-D embeddings which are then normalized to define a unitary hypersphere that represents each teacher's feature space. Both IResNet-34 and IResNet-50 backbones were tested, as this altered version of the ResNet architecture \cite{he2016deep} is a standard network in FR literature \cite{deng2018arcface, boutros2021elastic, neto2021focusface}. The normalized features are then processed by a classification header, a single fully connected layer that outputs a 7000-D logit vector. In classification tasks, it is common to transform this vector in class probabilities using the softmax loss function:

\begin{equation}
    L_{softmax}=-log\frac{e^{p_{y_i}}}{e^p_{y_i} + \sum_{j=1, j\neq y_i}^Ce^{p_j}},
    \label{eq:softmax}
\end{equation}
where $y_i$ is the ground truth label of sample $i$, $p_{y_i}$ is the value of the logit vector for class $y_i$ and $C$ is the number of classes. However, in FR tasks, this function is usually substituted by an altered version that focus on decreasing intra-class variability while increasing inter-class variability. In this work, the ElasticArcFace loss \cite{boutros2021elastic} was used for this purpose:

\begin{equation}
    L_{EAF}=-log\frac{e^{s(cos(p_{y_i}+E(m,\sigma)))}}{e^{s(cos(p_{y_i}+E(m,\sigma)))}+ \sum_{j=1, j\neq y_i}^Ce^{s(cos(p_j))}},
    \label{eq:ElasticArc}
\end{equation}
where $s$ and $m$ are the ElasticArcFace scale and margin, respectively, $\sigma$ is the standard deviation of the Gaussian distribution used to randomly modify the margin penalty values and the remaining variants are defined in the same way as in Equation \ref{eq:softmax}. In this work, the values assigned to $s$, $m$ and $\sigma$ were 64, 0.5 and 0.05, respectively, to retain the values recommended in the literature. 

The teacher for each ethnicity is retrieved from the best epoch for that ethnicity. These four teachers are then used to extract the embeddings of the original training samples, which will be fed to the adaptor network.

\subsection{Space Adaptors} \label{sec:space_adaptors}
\begin{figure}
    \centering
    \includegraphics[width=0.8\linewidth]{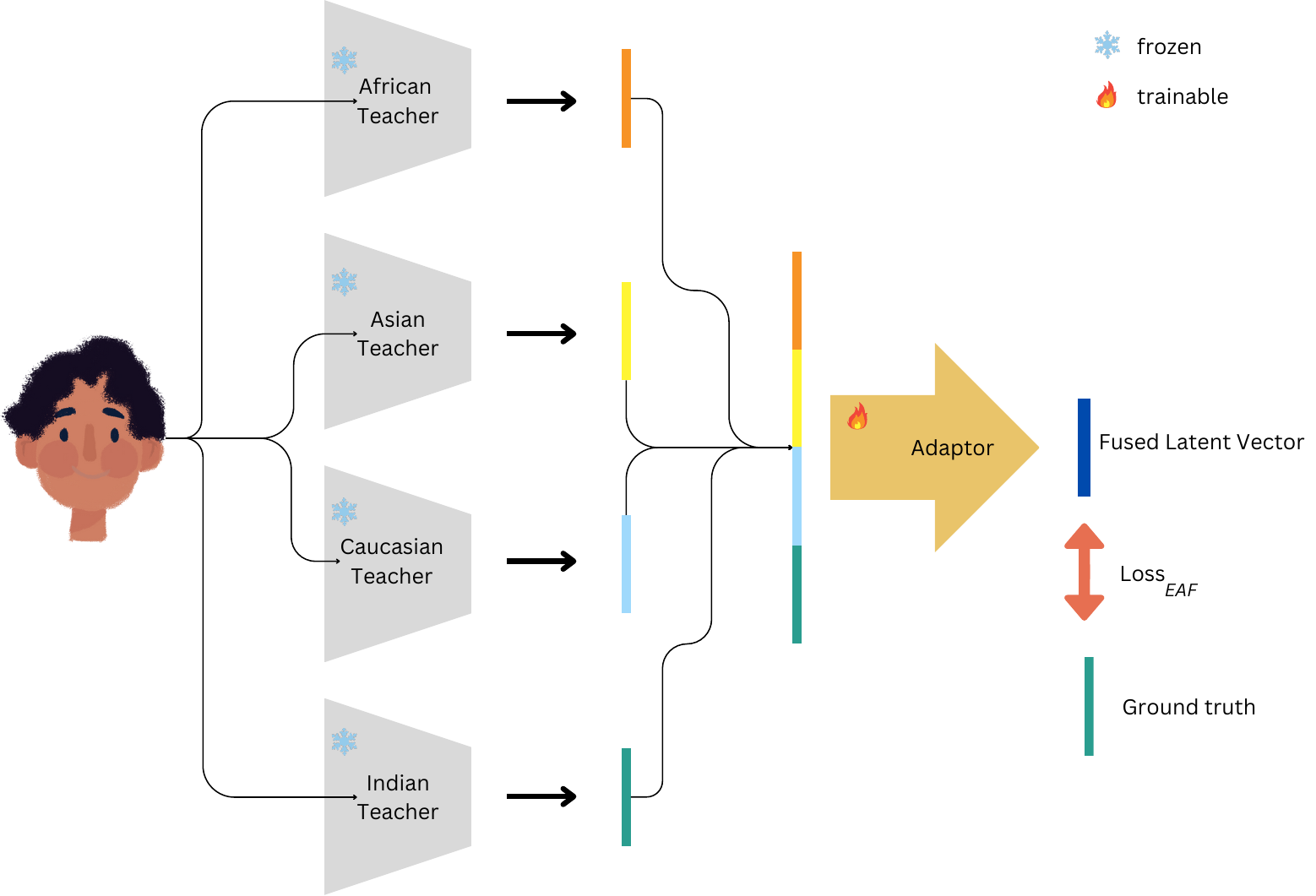}
    \caption{Training process of the proposed adaptor networks. Note that the previously trained teacher models are frozen during this process.}
    \label{fig:adaptor}
\end{figure}

Since knowledge is being transferred from more than one teacher, there is the need to adapt the four teachers' feature spaces to a common multi-teacher space from where knowledge can be directly distilled to the student. This can be achieved by training an adaptor network that simultaneously projects values from the four hyperspheres to a common one, as represented in Figure \ref{fig:adaptor}.

\vspace{-0.5cm}

The proposed adaptor networks contain a backbone and a header. The backbone is responsible for transforming the four teachers' spaces into a common 512-D space. To consider the information extracted by the four teachers simultaneously, their embeddings are concatenated to form the 2048-D vector that is inputted to this backbone. This vector contains four 512-D blocks corresponding to the features extracted by the teacher trained with samples of each ethnicity (Asian, African, Caucasian and Indian, respectively). This means that all teacher outputs are similarly weighted when the input vector is appended but their projection to the new hypersphere is learned. Furthermore, no information regarding each sample's ethnicity is needed to perform the space adaption, allowing for a privacy-friendly student train, especially when the classification term is dropped upon distillation \cite{shahreza2023synthdistill}, making this approach simple to deploy in real-world scenarios for which collecting demographic data might not be allowed due to privacy concerns. The header is a classification layer based on the ElasticArcFace loss \cite{boutros2021elastic} and can be ignored after training since the knowledge will be directly distilled from the multi-teacher space. As the considered loss function and the way the teacher embeddings are assembled are the most distinctive characteristics of these networks, they were named EAF-Fusion adaptors. In this work, we test three EAF-Fusion adaptors:

\begin{enumerate}
    \item \textbf{SL}: the backbone consists of a single fully connected layer that generates the 512-D multi-teacher space;
    \item \textbf{DuL}: the backbone consists of two fully connected layers. The first converts the 2048-D input vector into a 512-D embedding. A LeakyReLU \cite{maas2013rectifierNI} with a negative slope of 0.01 is used as the activation function. After activation, this vector is further processed by the second layer, generating the 512-D multi-teacher space;
    \item \textbf{DLDPO}: the backbone is similar to DuL's but dropout is applied with a probability of 0.2 before the LeakyReLU activation.
\end{enumerate}

The studied adaptors are trained on the complete BUPT-Balancedface dataset; the final adaptor is selected as the one presenting the lowest training loss. This network is then used to convert the teacher spaces into a common embedding space that will be used during the KD procedure. 

\rev{It should be noted that the proposed multi-teacher and adaptor framework makes our method flexible for different numbers of ethnicities, by adjusting the number of teachers and the adaptor input size. Nonetheless, it will lead to an increase in complexity and computation time, as more teachers need to be trained. On the other hand, increasing the samples per ethnicity is expected to boost our performance and fairness.}

\subsection{Students} \label{sec:students}

The knowledge of the multi-teacher space is integrated into the student's learning process by means of a KD loss, $L_{KD}$, as represented in Figure \ref{fig:students}. $L_{KD}$ minimizes the mean squared error (MSE) between the embeddings extracted by the student, $e_s$, and the multi-teacher space, $e_{mt}$:

\begin{figure}
    \centering
    \includegraphics[width=0.8\linewidth]{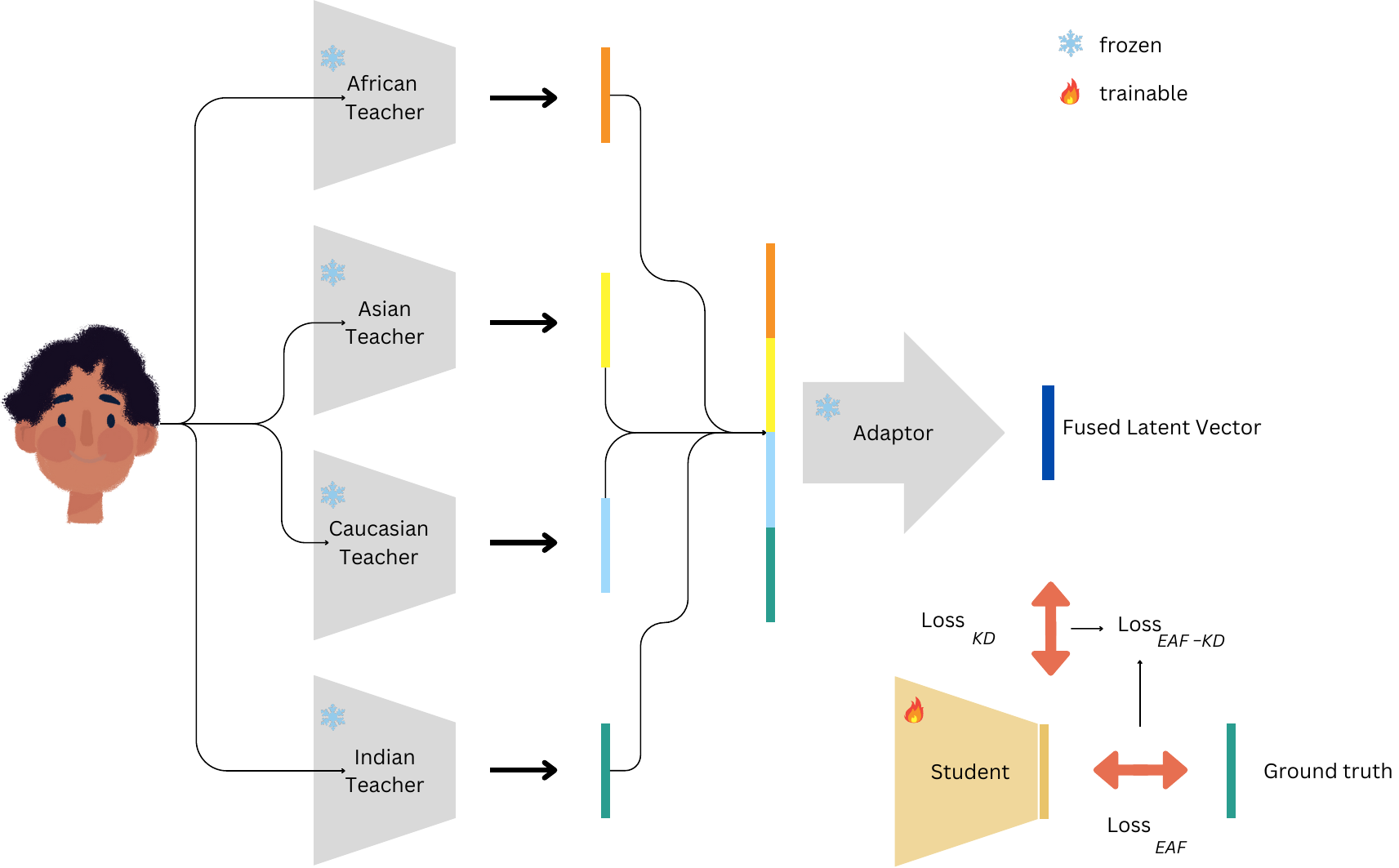}
    \caption{Training process of the proposed student networks using EAF-KD. For a-KD students, the classification loss term is disregarded; thus, no ground truth labels are required during training. Note that the previously trained teachers and adaptor are frozen during this process.}
    \label{fig:students}
\end{figure}

\begin{equation}
    L_{KD}=\frac{1}{D}\sum_{d=1}^D(e_{mt}-e_s)^2
    \label{eq:loss_KD},
\end{equation}
where $D=512$ is the number of dimensions of $e_{mt}$ and $e_s$. 

The student models assessed in this study fall into two groups: students trained with KD and the ElasticArcFace loss (EAF-KD) and students trained with KD alone (a-KD). EAF-KD students architecture is a IResNet-34 backbone responsible for extracting each sample's embedding, which is then normalized to generate the model's hypersphere. The normalized embeddings are then used by a header to perform classification, resulting in a 28000-D logits vector since the students are trained in the complete BUPT-Balancedface dataset. These logits are incorporated in the ElasticArcFace loss term given by Equation \ref{eq:ElasticArc}. The student loss in this scenario is defined as follows:

\begin{equation}
    \label{eq:loss_KDCEL}
    L_{EAF-KD}=L_{EAF}+\lambda L_{KD},
\end{equation}
where $\lambda$ is a hyperparameter that weights the importance of $L_{KD}$ for the total loss. This value was set empirically to 10000 to allow the model to learn from both terms simultaneously, after checking a predefined set of possible lambdas. 

The a-KD students only incorporate the IResNet-34 backbone responsible for extracting the 512-D embeddings which are then normalized. This type of procedure does not require access to the identity labels of the training data resulting in a more privacy-friendly setting \cite{shahreza2023synthdistill} whose performance should be evaluated. The final student loss is defined as:

\begin{equation}
    \label{eq:loss_KDalone}
    L_{a-KD}=\lambda L_{KD},
\end{equation}
where $\lambda$ is a hyperparameter used to control the magnitude of $L_{KD}$ and was set to 10000 to provide a framework comparable with the proposed EAF-KD. 

\subsection{Baseline} \label{sec:baseline}

To demonstrate that the effectiveness of the suggested approaches rely on the fact that information is being distilled from ethnicity-specialized teachers, a multi-teacher baseline was also trained and tested. The original identities from the BUPT-Balancedface dataset were randomly split into four disjoint subsets preserving data balance regarding ethnicity and each baseline teacher was trained in one of them, as displayed in Figure \ref{fig:data_split}. To keep consistency with the original teachers' training, each baseline teacher is assigned to one of the four considered ethnicities and its final version is selected as the model version that performs better for that ethnicity. Then, the baseline teachers knowledge is used to perform KD following the procedure defined in the previous sections. All the architectures and hyperparameters are the ones described for the original models to keep consistency and allow for direct comparison between them and the baseline.

\subsection{Experimental Setup} \label{sec:exp_setup}
All teacher models were trained for 52 epochs with a learning rate of 0.1 that decayed by a factor of 10 in epochs 16, 28, 40 and 50. The students were trained for 26 epochs with a learning rate of 0.1 that was  divided by 10 in epochs 8, 14, 20 and 25. The EAF-Fusion adaptors were trained on the complete BUPT-Balancedface dataset for 26 epochs with a learning rate of 1 that decayed by a factor of 10 in epochs 8, 14, 20 and 25. A batch size of 128 was considered and the models used a SGD optimizer with a momentum of 0.9. During all training processes, the tensors corresponding to each sample were normalized with $mean=[0.5, 0.5, 0.5]$ and $std=[0.5, 0.5, 0.5]$. Data augmentation was performed by flipping each image horizontally with a probability of 50\%.

To aid in the fairness analysis, the accuracy of each model was assessed separately for each ethnicity. The average of these four values was also computed to obtain a global performance metric. Regarding fairness evaluation, both the standard deviation between the four accuracy values (STD) and the skewed error ratio (SER) were determined for each architecture. 
By contrasting the worst and best classification mistakes, the SER facilitates comprehension of the relative differences in the accuracy associated with the four ethnicities:

\begin{equation}
    SER=\frac{100-min(acc)}{100-max(acc)},
\end{equation}
where $acc$ is a vector containing the four intra-ethnicity accuracy values.

\section{Results} \label{sec:results}
\subsection{Teachers} \label{sec:res_teachers}

The upper section of Table \ref{tab:teachers_34} displays the test results of the four IResNet-34 teachers trained on a single ethnicity. The best results for each metric are in bold. Each teacher achieves the best performance in identities from the ethnicity it was trained with, as expected. Furthermore, the bottom section of Table \ref{tab:teachers_34} shows that the baseline teachers achieved similar values for all evaluated metrics, proving that the data split performed before training is essential to obtain specialized teachers. This also means that the baseline teachers present reduced bias, as can be seen by the significantly lower STD and SER values. We aim to prove that distilling knowledge from teachers that are highly biased due to their ethnic specialization can lead to increased student fairness and performance.

T-Af outperformed the remaining specialized teachers in terms of global accuracy and fairness. This aligns with the fact that African identities are intrinsically harder to classify \cite{wang2019racial}, justifying the lower intra-ethnicity values achieved for the baseline teachers, which were trained on ethnicity-balanced data. 

\begin{table}[h!]
    \centering
    \tiny
    \caption{Performance and bias metrics of the four IResNet-34 teachers trained on a single ethnicity (\textbf{Ours}) and of the four IResNet-34 baseline teachers (\textbf{Baseline}) on the test set. The first four columns represent the intra-ethnicity accuracy levels. The remaining columns represent global performance and fairness metrics (STD and SER). The best results for each metric are marked in bold for each set of four teachers.}
    \label{tab:teachers_34}
    \begin{tabular}{|c|c||c|c|c|c||c|c|c|}
         \cline{3-9}
         \multicolumn{2}{c|}{} & \textbf{African} & \textbf{Asian} & \textbf{Caucasian} & \textbf{Indian} & \textbf{Global Acc} & \textbf{STD} & \textbf{SER} \\ 
         \hline
         \multirow{4}{*}{\textbf{Ours}} & \textbf{T-Af} & \textbf{89.82} & 78.32 & 86.87 & 86.00 & \textbf{85.25} & \textbf{4.90} & \textbf{2.13}\\
         & \textbf{T-As} & 70.85 & \textbf{89.63} & 81.10 & 79.85 & 80.36 & 7.69 & 2.81\\
         & \textbf{T-Ca} & 74.97 & 78.23	& \textbf{92.77} & 84.47 & 82.61 & 7.84 & 3.46\\
         & \textbf{T-In} & 73.45 & 78.22 & 83.68	& \textbf{89.18} & 81.13 & 6.80 & 2.45\\ 
         \hline \hline
         \multirow{4}{*}{\textbf{Baseline}} & \textbf{BT-1} & \textbf{85.38} & 87.02 & 90.42 & 88.37 & 87.80 & 2.13 & 1.53\\
         & \textbf{BT-2} & 85.37 & \textbf{87.30} & 90.40 & 88.23 & \textbf{87.83} & \textbf{2.09} & \textbf{1.52}\\
         & \textbf{BT-3} & 84.47 & 86.60 & \textbf{90.88} & 88.45 & 87.60 & 2.73 & 1.70\\ 
         & \textbf{BT-4} & 84.57 & 86.77	& 90.38 & \textbf{88.75} & 87.62 & 2.51 & 1.60\\ 
         \hline
    \end{tabular}
\end{table}



\subsection{a-KD}
The results achieved by the a-KD students trained with knowledge distilled from IResNet-34 teachers can be found in Table \ref{tab:akd_students_34}. It can be seen that students trained with knowledge extracted from specialized teachers consistently outperform their baseline versions in terms of global performance. Regarding bias, SL and DLDPO resulted in bias improvement when using specialized teachers. Although DuL achieved higher fairness when following the baseline protocol, it should be noted that our methodology surpasses the baseline in all intra-ethnicity accuracy metrics, meaning that our method still performs better than the baseline on ethnic minorities. Hence, SL can be considered the best method in terms of both performance and bias mitigation capabilities, which is interesting considering that no nonlinear behavior is incorporated in this strategy, resulting in a simple linear combination of the information provided by the teachers. Furthermore, these results indicate that the higher performances achieved by generalist teachers result in lower performance gains during distillation when compared with the specialized teachers, proving the usefulness of performing KD from specialized teachers for bias mitigation and performance improvement. 

\vspace{-0.5cm}

\begin{table}[h!]
    \centering
    \tiny
    \caption{RFW results of the a-KD students trained with knowledge distilled from IResNet-34 teachers. The best value achieved for each metric is highlighted in bold.}
    \label{tab:akd_students_34}
    \begin{tabular}{|c|c||c|c|c|c||c|c|c|}
         \cline{3-9}
         \multicolumn{2}{c|}{} & \textbf{African} & \textbf{Asian} & \textbf{Caucasian} & \textbf{Indian} & \textbf{Global Acc} & \textbf{STD} & \textbf{SER} \\ 
         \hline
      
   \multirow{3}{*}{\textbf{Ours}} & \textbf{SL} & 88.60 & \textbf{90.67} & 92.98 & \textbf{91.58} & \textbf{90.96} & 1.84 & 1.62\\	
         & \textbf{DuL} & \textbf{89.10} & 89.32 & \textbf{93.57} & 91.03 & 90.76 & 2.07 & 1.70\\
         & \textbf{DLDPO} & 88.02 & 89.35 & 92.18 & 89.85 & 89.85 & \textbf{1.73} & \textbf{1.53}\\
         \hline \hline
         \multirow{3}{*}{\textbf{Baseline}} & \textbf{SL} & \textbf{88.25} & \textbf{89.85} & \textbf{92.87} & \textbf{91.55} & \textbf{90.63} & 2.01 & 1.65\\	
         & \textbf{DuL} & 88.13 & 89.20 & 92.20 & 90.48 & 90.00 & \textbf{1.75} & 1.52\\
         & \textbf{DLDPO} & 87.55 & 88.42 & 91.63 & 89.58 & 89.30 & 1.76	& \textbf{1.49}\\
         \hline
    \end{tabular}
    
\end{table}

\subsection{EAF-KD} \label{sec:eafkd}
The results achieved by the EAF-KD students trained with knowledge distilled from IResNet-34 teachers can be found in Table \ref{tab:eaf_students_34}. Similarly to the previous approach, students trained with knowledge distilled from our teachers outperform the baseline. Once more, SL and DLDPO resulted in bias improvement when using specialized teachers, with SL reducing the STD from 1.54 to 1.36 while improving the overall performance by 0.42 percentual points.

\vspace{-0.5cm}

\begin{table}[h!]
    \centering
    \tiny
    \caption{RFW results of the EAF-KD students trained with knowledge distilled from IResNet-34 teachers. The best value achieved for each metric is highlighted in bold.}
    \label{tab:eaf_students_34}
    \begin{tabular}{|c|c||c|c|c|c||c|c|c|}
         \cline{3-9}
         \multicolumn{2}{c|}{} & \textbf{African} & \textbf{Asian} & \textbf{Caucasian} & \textbf{Indian} & \textbf{Global Acc} & \textbf{STD} & \textbf{SER} \\ 
         \hline
         \multirow{3}{*}{\textbf{Ours}} & \textbf{SL} & \textbf{92.12} & \textbf{93.07} & 95.33 & \textbf{93.93} & \textbf{93.61} & \textbf{1.36} & 1.69\\		
         & \textbf{DuL} & 91.80 & 91.33 & \textbf{95.42} & 92.68 & 92.81 & 1.83 & 1.89\\
         & \textbf{DLDPO} & 91.10 & 91.57 & 94.42 & 92.43 & 92.38 & 1.47 & \textbf{1.59}\\ 
         \hline \hline
         \multirow{3}{*}{\textbf{Baseline}} & \textbf{SL} & 91.43 & \textbf{92.68} & \textbf{95.10} & \textbf{93.53} & \textbf{93.19} & 1.54 & 1.75\\
         & \textbf{DuL} & \textbf{91.45} & 91.85 & 94.53 & 92.90 & 92.68 & \textbf{1.38} & \textbf{1.56}\\
         & \textbf{DLDPO} & 90.80 & 90.65 & 94.10 & 91.97 & 91.88 & 1.59 & 1.58\\ 
         \hline
    \end{tabular}
    
\end{table}

\vspace{-0.5cm}

As previously observed, more complex adaptors are keen to have lower performance than SL-EAF. This is an interesting behavior that indicates that a lightweight approach such as the linear combination of the teacher's knowledge is more effective, revealing that the teachers already hold the necessary information for the prediction and only a simple tuning function is needed to generate the multi-teacher space. This indicates that the proposed framework's bottleneck is probably associated with the teachers. Extra experiments performed to test this theory can be found in the supplementary material.

The joint analysis of Tables \ref{tab:teachers_34} and \ref{tab:eaf_students_34} shows that distilling knowledge from specialized teachers with a mean average accuracy of 82.33\% results in a student accuracy of 93.61\% while improving the average STD from 6.81 to 1.36 on SL. Hence, the proposed methodology allows for efficient conversion of four highly biased teachers into a single student model with better performance and fairness.

\section{Conclusion and Future Work} \label{sec:conclusion}

In this paper, we provide a novel multi-teacher KD strategy that is generally effective in reducing racial prejudice on the FR task while improving global performance. In the proposed framework, four specialized teacher models with the same architecture are trained separately on different ethnicities (African, Asian, Caucasian, and Indian). The four embedding spaces are then adapted to a common multi-teacher space using one of the three adaptor networks suggested in this paper. Each of these spaces is then utilized to distill knowledge into a student model using a-KD and EAF-KD. 

The obtained results show that utilizing specialized teachers to perform the distillation has a positive impact on model fairness while resulting in improved performance, proving the usefulness of the proposed methodology. Furthermore, the simplest adaptor architecture results in more efficient students, which suggests that an adaptor saturation results from the limited information provided by the teachers. Supported by the analysis provided in the supplementary material, we further concluded that this issue is not associated with the teachers' complexity level and is more likely to be attributed to the small dataset used to train these models, since during the four teachers' train each of them only has access to 7k distinct identities. In future work, these teachers should be trained in bigger datasets, namely synthetic datasets, to avoid privacy concerns.

\bibliographystyle{splncs04}
\bibliography{main}

\begin{thebibliography}{10}
\providecommand{\url}[1]{\texttt{#1}}
\providecommand{\urlprefix}{URL }
\providecommand{\doi}[1]{https://doi.org/#1}

\bibitem{DBLP:conf/bmvc/AlbieroB20}
Albiero, V., Bowyer, K.W.: Is face recognition sexist? no, gendered hairstyles and biology are. In: {BMVC} (2020)

\bibitem{DBLP:conf/icb/AlbieroZB20}
Albiero, V., Zhang, K., Bowyer, K.W.: How does gender balance in training data affect face recognition accuracy? In: {IJCB}. pp. 1--10. {IEEE} (2020)

\bibitem{aslam2023privileged}
Aslam, M.H., Zeeshan, M.O., Pedersoli, M., Koerich, A.L., Bacon, S., Granger, E.: Privileged knowledge distillation for dimensional emotion recognition in the wild. In: CVPRW. pp. 3337--3346 (2023)

\bibitem{blakeney2021simon}
Blakeney, C., Huish, N., Yan, Y., Zong, Z.: Simon says: Evaluating and mitigating bias in pruned neural networks with knowledge distillation. arXiv preprint arXiv:2106.07849  (2021)

\bibitem{boutros2021elastic}
Boutros, F., Damer, N., Kirchbuchner, F., Kuijper, A.: Elasticface: Elastic margin loss for deep face recognition. CoRR  \textbf{abs/2109.09416} (2021)

\bibitem{boutros2022template}
Boutros, F., Damer, N., Raja, K., Kirchbuchner, F., Kuijper, A.: Template-driven knowledge distillation for compact and accurate periocular biometrics deep-learning models. Sensors  \textbf{22}(5), ~1921 (2022)

\bibitem{boutros2022low}
Boutros, F., Kaehm, O., Fang, M., Kirchbuchner, F., Damer, N., Kuijper, A.: Low-resolution iris recognition via knowledge transfer. In: BIOSIG. pp.~1--5. IEEE (2022)

\bibitem{boutros2022pocketnet}
Boutros, F., Siebke, P., Klemt, M., Damer, N., Kirchbuchner, F., Kuijper, A.: Pocketnet: Extreme lightweight face recognition network using neural architecture search and multistep knowledge distillation. IEEE Access  \textbf{10},  46823--46833 (2022)

\bibitem{caldeira2023unveiling}
Caldeira, E., Neto, P.C., Gon{\c{c}}alves, T., Damer, N., Sequeira, A.F., Cardoso, J.S.: Unveiling the two-faced truth: Disentangling morphed identities for face morphing detection. In: {EUSIPCO}. pp. 955--959. {IEEE} (2023)

\bibitem{caldeira2024model}
Caldeira, E., Neto, P.C., Huber, M., Damer, N., Sequeira, A.F.: Model compression techniques in biometrics applications: A survey. arXiv preprint arXiv:2401.10139  (2024)

\bibitem{DBLP:conf/fgr/CaoSXPZ18}
Cao, Q., Shen, L., Xie, W., Parkhi, O.M., Zisserman, A.: Vggface2: {A} dataset for recognising faces across pose and age. In: 13th {IEEE} International Conference on Automatic Face {\&} Gesture Recognition, {FG} 2018, Xi'an, China, May 15-19, 2018. pp. 67--74. {IEEE} Computer Society (2018). \doi{10.1109/FG.2018.00020}, \url{https://doi.org/10.1109/FG.2018.00020}

\bibitem{DBLP:conf/icb/DebN018}
Deb, D., Nain, N., Jain, A.K.: Longitudinal study of child face recognition. In: {ICB}. pp. 225--232. {IEEE} (2018)

\bibitem{deng2018arcface}
Deng, J., Guo, J., Zafeiriou, S.: Arcface: Additive angular margin loss for deep face recognition. CoRR  \textbf{abs/1801.07698} (2018)

\bibitem{franco2021learn}
Franco, D., Oneto, L., Navarin, N., Anguita, D.: Learn and visually explain deep fair models: an application to face recognition. In: 2021 International Joint Conference on Neural Networks (IJCNN). pp. 1--10. IEEE (2021)

\bibitem{DBLP:conf/icb/FuD22}
Fu, B., Damer, N.: Towards explaining demographic bias through the eyes of face recognition models. In: {IJCB}. pp. 1--10. {IEEE} (2022)

\bibitem{fukuda2017efficient}
Fukuda, T., Suzuki, M., Kurata, G., Thomas, S., Cui, J., Ramabhadran, B.: Efficient knowledge distillation from an ensemble of teachers. In: Interspeech. pp. 3697--3701 (2017)

\bibitem{ge2018low}
Ge, S., Zhao, S., Li, C., Li, J.: Low-resolution face recognition in the wild via selective knowledge distillation. {IEEE} Trans. Image Process.  \textbf{28}(4),  2051--2062 (2018)

\bibitem{ge2020efficient}
Ge, S., Zhao, S., Li, C., Zhang, Y., Li, J.: Efficient low-resolution face recognition via bridge distillation. {IEEE} Trans. Image Process.  \textbf{29},  6898--6908 (2020)

\bibitem{gou2021knowledge}
Gou, J., Yu, B., Maybank, S.J., Tao, D.: Knowledge distillation: A survey. Int. J. Comput. Vis.  \textbf{129},  1789--1819 (2021)

\bibitem{guo2016ms}
Guo, Y., Zhang, L., Hu, Y., He, X., Gao, J.: Ms-celeb-1m: A dataset and benchmark for large-scale face recognition. In: Computer Vision--ECCV 2016: 14th European Conference, Amsterdam, The Netherlands, October 11-14, 2016, Proceedings, Part III 14. pp. 87--102. Springer (2016)

\bibitem{he2016deep}
He, K., Zhang, X., Ren, S., Sun, J.: Deep residual learning for image recognition. In: CVPR. pp. 770--778 (2016)

\bibitem{DBLP:journals/pami/HuangLLT20}
Huang, C., Li, Y., Loy, C.C., Tang, X.: Deep imbalanced learning for face recognition and attribute prediction. {IEEE} Trans. Pattern Anal. Mach. Intell.  \textbf{42}(11),  2781--2794 (2020)

\bibitem{huang2022evaluation}
Huang, Y., Wu, J., Xu, X., Ding, S.: Evaluation-oriented knowledge distillation for deep face recognition. In: CVPR. pp. 18740--18749 (2022)

\bibitem{huber2021mask}
Huber, M., Boutros, F., Kirchbuchner, F., Damer, N.: Mask-invariant face recognition through template-level knowledge distillation. In: FG. pp.~1--8. IEEE (2021)

\bibitem{karkkainen2021fairface}
Karkkainen, K., Joo, J.: Fairface: Face attribute dataset for balanced race, gender, and age for bias measurement and mitigation. In: WACV. pp. 1548--1558 (2021)

\bibitem{kolf2023syper}
Kolf, J.N., Elliesen, J., Boutros, F., Proen{\c{c}}a, H., Damer, N.: Syper: Synthetic periocular data for quantized light-weight recognition in the nir and visible domains. Image Vis. Comput.  \textbf{135},  104692 (2023)

\bibitem{DBLP:conf/iccvw/LiuZSYL21}
Liu, B., Zhang, S., Song, G., You, H., Liu, Y.: Rectifying the data bias in knowledge distillation. In: {ICCVW}. pp. 1477--1486. {IEEE} (2021)

\bibitem{liu2022coupleface}
Liu, J., Qin, H., Wu, Y., Guo, J., Liang, D., Xu, K.: Coupleface: Relation matters for face recognition distillation. In: ECCV. pp. 683--700. Springer (2022)

\bibitem{luo2016face}
Luo, P., Zhu, Z., Liu, Z., Wang, X., Tang, X.: Face model compression by distilling knowledge from neurons. In: AAAI. vol.~30 (2016)

\bibitem{maas2013rectifierNI}
Maas, A.L., Hannun, A.Y., Ng, A.Y., et~al.: Rectifier nonlinearities improve neural network acoustic models. In: Proc. icml. vol.~30, p.~3. Atlanta, GA (2013)

\bibitem{melo2023synthesis}
Melo, T., Cardoso, J., Carneiro, A., Campilho, A., Mendonca, A.M.: Oct image synthesis through deep generative models. In: CBMS. pp. 561--566 (2023). \doi{10.1109/CBMS58004.2023.00279}

\bibitem{neto2021focusface}
Neto, P.C., Boutros, F., Pinto, J.R., Damer, N., Sequeira, A.F., Cardoso, J.S.: Focusface: Multi-task contrastive learning for masked face recognition. In: 2021 16th IEEE International Conference on Automatic Face and Gesture Recognition (FG 2021). pp. 01--08. IEEE (2021)

\bibitem{neto2023compressed}
Neto, P.C., Caldeira, E., Cardoso, J.S., Sequeira, A.F.: Compressed models decompress race biases: What quantized models forget for fair face recognition. In: BIOSIG. pp.~1--5. IEEE (2023)

\bibitem{neto2024beyond}
Neto, P.C., Damer, N., Cardoso, J.S., Sequeira, A.F.: Beyond black and white: A more nuanced approach to facial recognition with continuous ethnicity  (2024), arxiv

\bibitem{neto2022explainable}
Neto, P.C., Gon{\c{c}}alves, T., Pinto, J.R., Silva, W., Sequeira, A.F., Ross, A., Cardoso, J.S.: Causality-inspired taxonomy for explainable artificial intelligence. arXiv preprint arXiv:2208.09500  (2024)

\bibitem{pham2023collaborative}
Pham, C., Hoang, T., Do, T.T.: Collaborative multi-teacher knowledge distillation for learning low bit-width deep neural networks. In: Proceedings of the IEEE/CVF Winter Conference on Applications of Computer Vision. pp. 6435--6443 (2023)

\bibitem{innocenceproject2024race}
Project, I.: Race and wrongful convictions (2024), available: \url{https://innocenceproject.org/race-and-wrongful-conviction/}. Accessed: 2024-06-09

\bibitem{robinson2020face}
Robinson, J.P., Livitz, G., Henon, Y., Qin, C., Fu, Y., Timoner, S.: Face recognition: too bias, or not too bias? In: CVPRW. pp.~0--1 (2020)

\bibitem{shahreza2023synthdistill}
Shahreza, H.O., George, A., Marcel, S.: Synthdistill: Face recognition with knowledge distillation from synthetic data. In: 2023 IEEE International Joint Conference on Biometrics (IJCB). pp. 1--10. IEEE (2023)

\bibitem{terhorst2020compr}
Terh{\"o}rst, P., Kolf, J.N., Huber, M., Kirchbuchner, F., Damer, N., Moreno, A.M., Fierrez, J., Kuijper, A.: A comprehensive study on face recognition biases beyond demographics. IEEE Transactions on Technology and Society  \textbf{3}(1),  16--30 (2021)

\bibitem{wang2019racial}
Wang, M., Deng, W., Hu, J., Tao, X., Huang, Y.: Racial faces in the wild: Reducing racial bias by information maximization adaptation network. In: ICCV. pp. 692--702 (2019)

\bibitem{wang2021meta}
Wang, M., Zhang, Y., Deng, W.: Meta balanced network for fair face recognition. {IEEE} Trans. Pattern Anal. Mach. Intell.  \textbf{44}(11),  8433--8448 (2021)

\bibitem{wang2021teacher}
Wang, X.: Teacher guided neural architecture search for face recognition. In: AAAI. vol.~35, pp. 2817--2825 (2021)

\bibitem{wu2019multi}
Wu, M.C., Chiu, C.T., Wu, K.H.: Multi-teacher knowledge distillation for compressed video action recognition on deep neural networks. In: ICASSP 2019-2019 IEEE International Conference on Acoustics, Speech and Signal Processing (ICASSP). pp. 2202--2206. IEEE (2019)

\bibitem{wu2020learning}
Wu, X., He, R., Hu, Y., Sun, Z.: Learning an evolutionary embedding via massive knowledge distillation. International Journal of Computer Vision  \textbf{128},  2089--2106 (2020)

\bibitem{xu2023probabilistic}
Xu, J., Li, S., Deng, A., Xiong, M., Wu, J., Wu, J., Ding, S., Hooi, B.: Probabilistic knowledge distillation of face ensembles. In: CVPR. pp. 3489--3498 (2023)

\bibitem{xu2020investigating}
Xu, T., White, J., Kalkan, S., Gunes, H.: Investigating bias and fairness in facial expression recognition. In: ECCVW. pp. 506--523. Springer (2020)

\bibitem{yi2014learning}
Yi, D., Lei, Z., Liao, S., Li, S.Z.: Learning face representation from scratch. arXiv preprint arXiv:1411.7923  (2014)

\bibitem{you2017learning}
You, S., Xu, C., Xu, C., Tao, D.: Learning from multiple teacher networks. In: Proceedings of the 23rd ACM SIGKDD international conference on knowledge discovery and data mining. pp. 1285--1294 (2017)

\bibitem{zhang2022confidence}
Zhang, H., Chen, D., Wang, C.: Confidence-aware multi-teacher knowledge distillation. In: ICASSP 2022-2022 IEEE International Conference on Acoustics, Speech and Signal Processing (ICASSP). pp. 4498--4502. IEEE (2022)

\bibitem{zhao2023grouped}
Zhao, W., Zhu, X., Guo, K., Zhang, X., Lei, Z.: Grouped knowledge distillation for deep face recognition pp. 3615--3623 (2023)

\bibitem{zoph2016neural}
Zoph, B., Le, Q.: Neural architecture search with reinforcement learning. In: ICLR (2017), \url{https://openreview.net/forum?id=r1Ue8Hcxg}

\end{thebibliography}

\newpage
\section*{Supplementary Material} \label{sec:sup_mat}
As concluded in Section 5.3, the simplest adaptor function, SL, is the more effective one, suggesting that only a simple training function is needed to generate an appropriate multi-teacher space from which knowledge can be distilled. This indicates that the proposed framework's bottleneck is probably associated with the teachers. As the amount of information extracted by these networks might not be enough to prevent adaptor saturation, we increase the teachers' complexity to an IResNet-50 and analyze the obtained results in this supplementary material.

\subsection*{Teachers} \label{sec:res_teachers_50}
The upper section of Table \ref{tab:teachers_50} displays the test results of the four IResNet-50 teachers trained on a single ethnicity. In line with what was shown in Section 5.1, each teacher achieves the best performance in identities from the ethnicity it was trained with and the baseline teachers achieved similar values for all evaluated metrics. This reinforces the previously drawn conclusion regarding the importance of the data split performed before training to obtain specialized teachers. Once more, T-Af outperformed the remaining specialized teachers in terms of global accuracy and fairness.

As expected, the IResNet-50 teachers surpassed their correspondent IResNet-34 versions (Table 1) in terms of global performance, suggesting that using these teachers might result in improved students' performance and fairness.

\begin{table}[]
    \centering
    \scriptsize
    \begin{tabular}{|c|c||c|c|c|c||c|c|c|}
         \cline{3-9}
         \multicolumn{2}{c|}{} & \textbf{African} & \textbf{Asian} & \textbf{Caucasian} & \textbf{Indian} & \textbf{Global Acc} & \textbf{STD} & \textbf{SER} \\ 
         \hline
         \multirow{4}{*}{\textbf{Ours}} & \textbf{T-Af} & \textbf{90.07} & 78.88 & 87.02 & 86.80 & \textbf{85.69} & \textbf{4.78} & \textbf{2.13}\\
         & \textbf{T-As} & 71.97 & \textbf{90.33} & 82.15 & 80.95 & 81.35 & 7.51 & 2.90\\
         & \textbf{T-Ca} & 76.62 & 79.38 & \textbf{92.67} & 84.68 & 83.34 & 7.06 & 3.19\\
         & \textbf{T-In} & 75.25 & 78.73 & 83.90 & \textbf{88.60} & 81.62 & 5.86 & 2.17\\ 
         \hline \hline
         \multirow{4}{*}{\textbf{Baseline}} & \textbf{BT-1} & 85.73 & 87.70 & \textbf{90.95} & 88.83 & \textbf{88.30} & 2.18 & 1.58\\
         & \textbf{BT-2} & \textbf{85.92} & 87.55 & 90.35 & 88.17 & 88.00 & \textbf{1.83} & \textbf{1.46}\\
         & \textbf{BT-3} & 85.55 & 87.03 & 90.93	& \textbf{88.92} & 88.11 & 2.33 & 1.59\\ 
         & \textbf{BT-4} & 85.28	& \textbf{87.77} & 90.62 & 88.48	& 88.04 & 2.20 & 1.57\\ 
         \hline
    \end{tabular}
    \caption{Performance and bias metrics of the four IResNet-50 teachers trained on a single ethnicity (\textbf{Ours}) and of the four IResNet-50 baseline teachers (\textbf{Baseline}) on the test set. The best results per metric are marked in bold for each set of four teachers.}
    \label{tab:teachers_50}
\end{table}

\subsection*{Students}
The results achieved by the a-KD and EAF-KD students trained with knowledge distilled from IResNet-34 teachers can be found in Tables \ref{tab:akd_students_50} and \ref{tab:eaf_students_50}, respectively. The comparison between these results and the ones achieved by the students trained with knowledge distilled from IResNet-34 teaches (Tables 2 and 3) shows that no significant improvements were verified when distilling knowledge from a more complex network. In some scenarios, the students originated from less complex teachers even surpassed those trained with IResNet-50 information. This shows that the adaptor saturation issue might be associated with teacher-related factors that are not related with these networks complexity. For example, this phenomenon may arise from the low amount of information used to train each teacher, as these models are only fed 7k distinct identities.

\begin{table}[]
    \centering
    \scriptsize
    \begin{tabular}{|c|c||c|c|c|c||c|c|c|}
         \cline{3-9}
         \multicolumn{2}{c|}{} & \textbf{African} & \textbf{Asian} & \textbf{Caucasian} & \textbf{Indian} & \textbf{Global Acc} & \textbf{STD} & \textbf{SER} \\ 
         \hline
         \multirow{3}{*}{\textbf{Ours}} & \textbf{SL} & 89.52 & 89.68 & 92.95 & \textbf{91.63} & 90.95 & 1.65 & \textbf{1.49} \\	
         & \textbf{DuL} & \textbf{90.08} & \textbf{90.13} & \textbf{93.48} & 90.88 &	\textbf{91.14} & \textbf{1.60} & 1.52\\
         & \textbf{DLDPO} & 88.17 & 89.62 & 92.42 & 90.32 & 90.13 & 1.77 & 1.56\\
         \hline \hline
         \multirow{3}{*}{\textbf{Baseline}} & \textbf{SL} & 88.40 & \textbf{90.53} & \textbf{93.42} & \textbf{91.33} & \textbf{90.92} & 2.08 & 1.76\\	
         & \textbf{DuL} & \textbf{88.45} & 89.92 & 92.73 & 90.90 & 90.50 & 1.80 & 1.59\\
         & \textbf{DLDPO} & 87.43 & 88.62 & 91.65 & 89.50 & 89.30 & \textbf{1.78} & \textbf{1.51}\\
         \hline
    \end{tabular}
    \caption{Test results of the a-KD students trained with knowledge distilled from IResNet-50 teachers. The best value achieved for each metric is highlighted in bold.}
    \label{tab:akd_students_50}
\end{table}

\begin{table}[]
    \centering
    \scriptsize
    \begin{tabular}{|c|c||c|c|c|c||c|c|c|}
         \cline{3-9}
         \multicolumn{2}{c|}{} & \textbf{African} & \textbf{Asian} & \textbf{Caucasian} & \textbf{Indian} & \textbf{Global Acc} & \textbf{STD} & \textbf{SER} \\ 
         \hline
         \multirow{3}{*}{\textbf{Ours}} & \textbf{SL} & \textbf{92.30} & \textbf{92.88} & \textbf{95.28} & \textbf{93.52} & \textbf{93.50} & \textbf{1.29} & 1.63\\		
         & \textbf{DuL} & 91.42 & 91.43 & \textbf{95.28} & 92.47 & 92.65 & 1.82 & 1.82\\
         & \textbf{DLDPO} & 91.25 & 91.71 & 94.55 & 92.53 & 92.51 & 1.46 & \textbf{1.61}\\ 
         \hline \hline
         \multirow{3}{*}{\textbf{Baseline}} & \textbf{SL} & \textbf{91.77} & \textbf{92.37} & \textbf{94.92} & \textbf{93.00} & \textbf{93.02} & \textbf{1.37} & 1.62 \\
         & \textbf{DuL} & 91.72 & 91.97 & 94.82 & 92.75 & 92.82 & 1.41 & 1.60\\
         & \textbf{DLDPO} & 90.42 & 90.93 & 93.67 & 91.58 & 91.65 & 1.43	& \textbf{1.51}\\ 
         \hline
    \end{tabular}
    \caption{Test results of the EAF-KD students trained with knowledge distilled from IResNet-50 teachers. The best value achieved for each metric is highlighted in bold.}
    \label{tab:eaf_students_50}
\end{table}
\end{document}